\documentclass[10pt, a4paper]{article}
\usepackage{CJKutf8}
\usepackage{booktabs}
\usepackage{bm}
\usepackage{multirow}
\usepackage{amsmath}
\usepackage{lrec-coling2024} 
\title{Robust Guidance for Unsupervised Data Selection:\\Capturing Perplexing Named Entities\\for Domain-Specific Machine Translation}

\name{Seunghyun Ji$^{12}$, Hagai Raja Sinulingga$^2$, Darongsae Kwon$^2$} 

\address{$^1$Ahancompany corporation, $^2$TelePIX \\
         seunghyun.ji@a-ha.io, \{hagairaja, darong.kwon\}@telepix.net}
         
\abstract{
Low-resourced data presents a significant challenge for neural machine translation. In most cases, the low-resourced environment is caused by high costs due to the need for domain experts or the lack of language experts. Therefore, identifying the most training-efficient data within an unsupervised setting emerges as a practical strategy. Recent research suggests that such effective data can be identified by selecting 'appropriately complex data' based on its volume, providing strong intuition for unsupervised data selection. However, we have discovered that establishing criteria for unsupervised data selection remains a challenge, as the 'appropriate level of difficulty' may vary depending on the data domain. We introduce a novel unsupervised data selection method named 'Capturing Perplexing Named Entities,' which leverages the maximum inference entropy in translated named entities as a metric for selection. When tested with the 'Korean-English Parallel Corpus of Specialized Domains,' our method served as robust guidance for identifying training-efficient data across different domains, in contrast to existing methods.
\\ \newline   }

\begin{document}

\maketitleabstract

\section{Introduction}
{\let\thefootnote\relax\footnotetext{This work was initially started in TelePIX, the previous affiliation of the first author.}}
{\let\thefootnote\relax\footnotetext{The code is available in the following hyperlink : \url{https://github.com/comchobo/Capturing-Perplexing-Named-Entities}}}

With the advent of large-scale models capable of translating numerous languages in various directions\citep{aharoni-etal-2019-massively}, the field of machine translation is entering a new era. For instance, 'No Language Left Behind\citep{nllb2022}', which demonstrated outstanding performance across a range of languages, was trained on over 40,000 combinations of 200 languages. These models can be regarded as pre-trained or foundational, as they have acquired general knowledge for translation. Nevertheless, they might sometimes face challenges when translating domain-specific data, despite their extensive training on diverse datasets. To address this, fine-tuning the pre-trained models with target domain data can enhance their specialization\cite{fadaee-monz-2018-back, zanvega}.

\begin{figure*}[t]
    \begin{center}
        \includegraphics[scale=0.5]{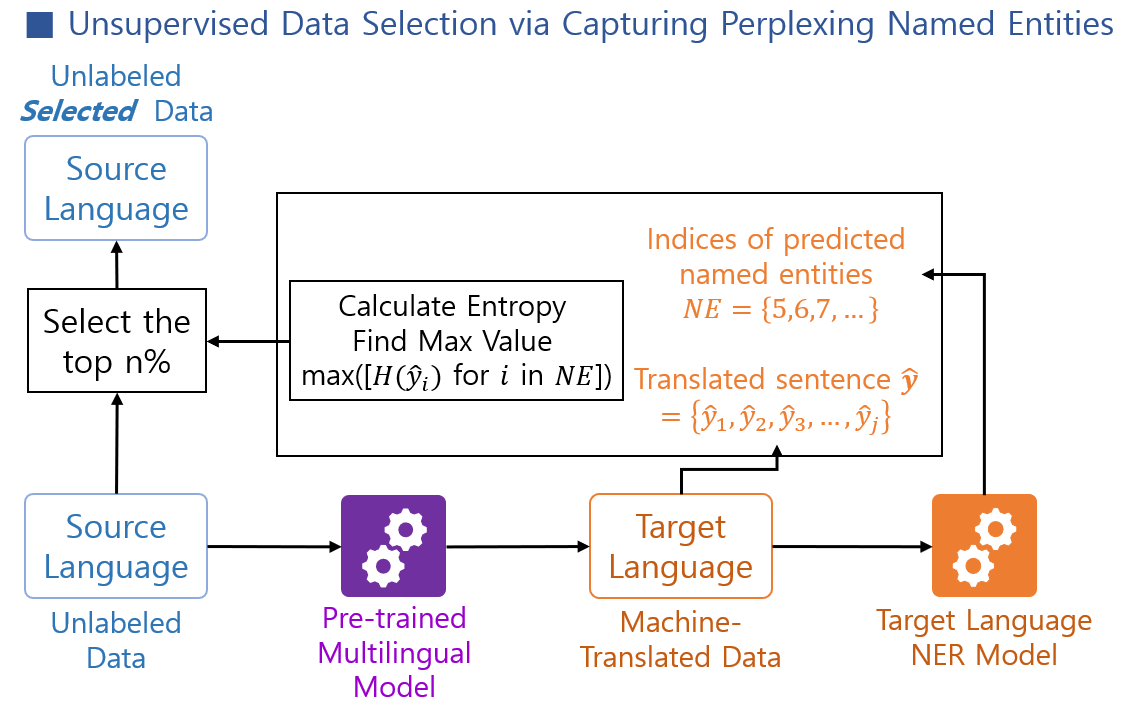} 
        \caption{A diagram illustrates our method, which utilizes a pre-trained multilingual model for machine translation and a named entity recognition model that has been fine-tuned on the target language. Our method comprises three steps: 1) capturing named entity tokens in the machine-translated sentences, 2) calculating the inference entropy of those tokens, and 3) using the maximum entropy value as a measure for selection.}
    \end{center}
    \label{fig:diagram}
\end{figure*}

However, when addressing narrow or specialized domains, the model must recognize words that are relatively rare in general corpora. This presents a challenge, as rare words often consist of sparse tokens, such as those composed of single character tokens. Named entities, such as names of persons, organizations, etc., frequently lack synonyms, making it even more perplexing to build contextualized representations, especially in narrow domains. This also underscores the point that acquiring domain-specific translation data is costly, as translators are required who possess not only domain expertise but also familiarity with domain-specific terminology.

To reduce data acquisition costs, one might consider strategically identifying data for labeling rather than making random selections. Several researchers\citep{paul2021deep, feldman2020neural, sorscher2022beyond} have suggested various measurement methods aimed at selecting 'effective' data for training. Some of those focus on 'Data difficulty,'\citep{paul2021deep, meding2022trivial} identifying data that poses a challenge to a given model. 'Data forgettability\citep{toneva2018an}' or 'Memorization\citep{feldman2020neural}' could serve as alternative criterion. However, these methods require a supervised setting for selection, which may be inefficient for machine translation. For instance, pruning a dataset is unlikely to yield a better model if the dataset was curated by domain experts \citep{maillard-etal-2023-small}.

In an unsupervised setting, where training-efficiency should be guessed without a label, \citet{sorscher2022beyond} demonstrated that the Euclidean distance between a data point's representation and its cluster centroid can serve as an effective criterion for data selection. This approach is supported by several concrete theoretical analyses and provides straightforward guidance for data selection. However, it remains uncertain whether this criterion can be universally applied to parameter-efficient fine-tuning methods\citep{houlsby2019parameter, hu2022lora, liu2022few}, which are commonly used. We observed that this measurement method might not always align with training-efficiency, indicating that it may not consistently correlate with performance improvement, despite using the same pre-trained weights and dataset size. These findings are detailed in Section \ref{sec:results}.

We propose a novel method for unsupervised data selection, which we refer to as 'Capturing Perplexing Named Entities'. Our method identifies data that should be selected, by assessing the perplexity of named entity tokens translated by a given pre-trained model, as described in Figure \ref{fig:diagram}. The motivations behind this approach are as follow:

\begin{itemize}
    \item{Since named entities in domain-specific data are challenging to translate without recognizing the complex patterns within the domain, they represent one of the most difficult portions to translate. Therefore, these entities should be given priority for efficient domain adaptation.}
    \item{The entropy score of a vocabulary distribution can indicate the model's level of perplexity. Given that synonyms for named entities are unlikely to exist, the model should not exhibit a high entropy score for named entities.}
\end{itemize}

In several experiments targeting domain-specific 'Korean to English' translation, our method consistently identified the most training-efficient data. This indicates that our measurement method has a stronger correlation with performance improvement compared to existing methods, which can vary significantly across different data domains. For clarity in our discussion, 'MDS' will serve as the abbreviation for Measurement method for Data Selection, and 'Value by MDS' will denote the specific value it calculates.

\section{Related Works}

\begin{table*}[!ht]
\begin{center}
\begin{tabularx}{\linewidth}{c|X|c}
    \toprule
    Languages & \centering{Data Examples} & Scores \\
    
    \midrule
    \multirow{3}{*}{Korean}&
    \begin{CJK}{UTF8}{mj}
        \multirow{3}{*}{\parbox{11.5cm}{메이저리그 자유계약선수(FA) 최대어 투수 중 한 명인 스티븐 스트라스버그가 원 소속팀 워싱턴과 7년 2억4,500만달러에 도장을 찍었다.}}
    \end{CJK} 
    & \multirow{2}{*}{COMET} \\
    & & \multirow{2}{*}{90.92} \\[1.5em]
    
    \multirow{3}{*}{English} &
    \multirow{3}{*}{\parbox{11.5cm}{\textcolor{red}{Steven Strasburg}, one of the biggest \textcolor{red}{free agent (FA) pitchers} in Major League Baseball, has signed a 7-year, \$ 245 million contracts with his \textcolor{red}{original} team Washington.}} 
    & \multirow{2}{*}{ChrF++} \\
    & & \multirow{2}{*}{67.94} \\[1.5em]
    
    \multirow{3}{*}{Translated}&
    \multirow{3}{*}{\parbox{11.5cm}{\textcolor{red}{Steven Strasberg}, one of the biggest pitchers in the Major League \textcolor{red}{Free Agent (FA) league}, signed a seven-year, \$ 245 million contract with \textcolor{red}{former} team Washington.}}
    & \multirow{2}{*}{BLEU} \\
    
    & & \multirow{2}{*}{27.38} \\[2em]
    
    \midrule

    \multirow{2}{*}{Korean}&
    \begin{CJK}{UTF8}{mj}
        \multirow{2}{*}{\parbox{11.5cm}{고메스 부상 이후 에버턴 지휘봉을 잡게된 카를로 안첼로티 감독은 지난주 "고메스의 회복이 순조롭게 이뤄지고 있다"고 밝혔다.}}
    \end{CJK} 
    & COMET\\
    & & 90.99 \\
    & & \\
    
    \multirow{2}{*}{English}&
    \multirow{2}{*}{\parbox{11.5cm}{\textcolor{red}{Manager} Carlo Ancelotti, who took the helm of Everton after \textcolor{red}{Gomez's} injury, revealed last week that \textcolor{red}{"Gomez's} recovery is going smoothly."}} 
    & ChrF++\\
    & & 64.47\\
    & & \\
    \multirow{2}{*}{Translated}&
    \multirow{2}{*}{\parbox{11.5cm}{\textcolor{red}{Coach} Carlo Ancelotti, who took over Everton after \textcolor{red}{Gomes'} injury, said last week, \textcolor{red}{"Gomes'} recovery is progressing smoothly."}} 
    & BLEU\\
    & & 24.86 \\
    & & \\
    \bottomrule
    
\end{tabularx}
\caption{Example pairs with high COMET and ChrF++ scores but low BLEU scores were selected from sports domain data. The first column represents the source (Korean), the target (English), and the machine-translated (Korean to English) result. Words that may cause critical semantic distortions are highlighted in red. The last column lists the evaluation scores of the machine-translated sentences, calculated using three different metrics.}
\label{table:comet-discrepancy}
 \end{center}
\end{table*}

\subsection{Named Entities in Machine Translation} \label{NEMT}

Translating named entities presents a significant challenge in machine translation\citep{ugawaetal2018neural}, although it is crucial for delivering accurate information\citep{tjong-kim-sang-de-meulder-2003-introduction}. Incorrect translations of named entities, even with few errors, can lead to information distortion. For instance, in Table \ref{table:comet-discrepancy}, the human-translated and machine-translated Korean to English-sentences may seem similar. However, a closer examination reveals differences in the individual's name (Steven Strasburg), the league (Major League Baseball), and an adjective (original). Despite these mistakes causing critical distortions, recent metrics such as COMET\citep{rei-etal-2020-comet}\footnote{We used \url{https://huggingface.co/Unbabel/wmt22-comet-da} to evaluate using COMET score.} and ChrF++\citep{popovic-2015-chrf} show scores high enough to be interpreted as satisfactory results. Given that some rare named entities are more common in domain-specific data, building precise contextualized representations of data, which contains named entities, is even difficult to capture by recent deep-model based metrics.

One current approach to translate named entities precisely, integrates a knowledge base\citep{zhao2020knowledge} or employs a transliteration model once tokens are identified as named entities\citep{10111938}. However, these strategies often rely on specialized algorithms that act as a workaround, rather than directly boosting the translation model's performance or robustness. Multi-task learning has demonstrated improvements in translation performance when additional annotations for named entities are provided\citep{xie2022end}. However, this method may incur significantly higher labeling costs.

\subsection{Data Selection for Training}

Throughout several training cycles, metrics such as forgetting scores\citep{toneva2018an}, memorization\citep{feldman2020neural}, diverse ensembles\citep{meding2022trivial}, and normed gradients\citep{paul2021deep} could be used as one of the measurement methods for data selection (MDS). EL2N, which quantifies the error magnitude, acts as a training-free MDS. However, these methods require annotations, limiting their application to supervised settings only. As high-quality data has been shown to significantly outperform large volumes of low-quality or synthetic data\citep{maillard-etal-2023-small}, it is generally recommended that the data with elaborate annotations should not be pruned.

In an unsupervised setting, one might explore data uniqueness—for example, by measuring the Euclidean distance between a data representation and its centroid\citep{sorscher2022beyond} (referred to as Selfsup)—as a form of unsupervised MDS. Measuring uncertainty, which could be estimated by the entropy of the probability distribution, also might be one of MDS\citep{brown1990statistical, wu2021uncertainty}. However, empirical evidence suggests that when training with small datasets, excessively unique data (indicated by high values in MDS Selfsup) may impede training\citep{sorscher2022beyond}. Therefore, selecting data using the appropriate type of MDS and determining the optimal value for MDS are crucial. Nonetheless, establishing a standard for this is challenging, to the best of our knowledge.

In machine translation, reference-free Quality Estimation (QE) methods, which operate as an unsupervised MDS, are  gaining focus. One strategy involves the intuition of 'seeking perplexing data' by identifying attention distractions or uncertainties\citep{peris-casacuberta-2018-active}. More sophisticated reference-free QE algorithms, which can be implemented using deep models\citep{rei-etal-2021-references}, have demonstrated competitive results when compared to their reference-requiring counterparts\citep{rei-etal-2020-comet}. However, these methods, relying on sentence embedding models, are often confounded by even slight literal differences. We have observed and discussed this phenomenon in Section \ref{NEMT}.

\section{Existing Methods}\label{sec:methods}
We consider the multilingual translation model as a 'pre-trained model', with subsequent training on specific data referred to as 'fine-tuning'. 

\subsection{EL2N}

\citet{paul2021deep} previously used the average error from several minimally trained models to identify data that could not be easily trained in a few epochs. This method requires paired data for its computations, hence categorized as a supervised approach. Intuitively, the EL2N value from a pre-trained model signifies an average error or incorrect confidence, enabling the identification of the most problematic data for a given model. If $\bm{Y}$ and $\bm{\hat{Y}}$ represent the original and translated sentences in the target language, respectively, EL2N can be described as follows: 

\begin{center}
    EL2N$(\bm{Y}, \bm{\hat{Y}})=
    \frac1L \sum_{l=1}^{L}
    \|\bm{y}_l-\hat{\bm{y}_l}\|$
    $L=min(|\bm{Y}|, |\bm{\hat{Y}}|)$
\end{center}
where $\bm{\hat{y}}$ represents the predicted token distribution, and $\bm{y}$ is the actual label. Given that the translated sentence may contain a different number of tokens from original sentence, we chose the shorter token length, represented by the cardinality of $\bm{Y}$ and $\bm{\hat{Y}}$.

\subsection{Entropy}
\citet{brown1990statistical} demonstrated that uncertainty in prediction is quantifiable by entropy. Various studies have reported performance improvements by employing entropy to select data for training\citep{jiao2021self, wu2021uncertainty}. Building on this concept, we considered entropy as an indicator of the pre-trained model's perplexity regarding specific sentences, selecting them as candidates for fine-tuning. The entropy of the vocabulary distribution is defined as:

\begin{center}
    $H(\bm{\hat{y}})
    =\frac1V
    \sum_{i\in{}V}-P(\hat{y_i})logP(\hat{y_i})$
\end{center}
where $V$ is a vocabulary. We adopted averaged entropy as MDS which is as follows:
\begin{center}
    $AvgEntropy(\bm{\hat{Y}})=
    \frac1L\sum_{l=1}^{L}H(\bm{\hat{y}}_l)$
\end{center}
where $L$ is a length of the sentence $\bm{\hat{Y}}$.

However, given that the optimal entropy level may differ by token types, such as adjectives or synonyms, we hypothesized that employing $AvgEntropy$ as an MDS might lead the model to become either overconfident or overly cautious.

\subsection{Selfsup}\label{subsec:selfsup}
\citet{sorscher2022beyond} observed that within clustered image representations, data points distant from their centroids often exhibit unique patterns, which have high Euclidean distance to the centroid. However, its effectiveness as an MDS for fine-tuning translation models remains unverified. To adapt this approach to the language domain, we utilized sentence embeddings for the source data and applied k-means clustering. If $x_A$ represents a sentence embedding of source language data $x$, clustered around centroid $A$, then the MDS Selfsup can be described as:

\begin{center}
    $Selfsup(x_A) = ||x_A - A||$
\end{center}

If the sentence embeddings are well-aligned, MDS Selfsup is expected to capture training-efficient data for fine-tuning. Although recent sentence embedding models demonstrate decent performance, their accuracy in domain-specific data remains questionable. Our findings provide support for this doubt, as illustrated in Table \ref{table:comet-discrepancy}, where the COMET score failed to detect semantic distortion.

\subsection{Reference-free COMET}
\citet{rei-etal-2021-references} proposed a Reference-free COMET, which was trained to estimate quality without reference, only with source and translated sentences. Reference-free COMET was designed to predict quality annotations using a sentence embedding model. Its output range is 0 to 1, where 1 denotes the best quality. We expected that Reference-free COMET as an MDS would be inversely proportional to the training-efficiency since it would detect examples that the model could not translate well.

\section{Proposed Method}

Our hypothesis posits that complex patterns possessed by named entities are essential for fine-tuning. This is particularly true in domain-specific machine translation, where rare words and expressions occur frequently but are not present in the general domain. By incorporating these characteristics into data selection, we measured the maximum entropy while translating named entities, which are unlikely to have alternative answers. In summary, our method specifically targets perplexing named entities.

\begin{center}
    $PerEnts(\bm{\hat{Y}}) = max(\{H(\bm{\hat{y}}_x) | x \in NE(\bm{\hat{y}})\})$\\
\end{center}
where $NE(\bm{\hat{y}})$ represents a set of named entity token indices in the machine-translated sentence $\bm{\hat{y}}$, predicted by a named entity recognition model. We will use the abbreviation 'PerEnts,' to refer to our method.

\section{Experiments}

\subsection{Settings for Experiments}

\begin{figure}[t]
    \begin{center}
        \includegraphics[width=\linewidth]{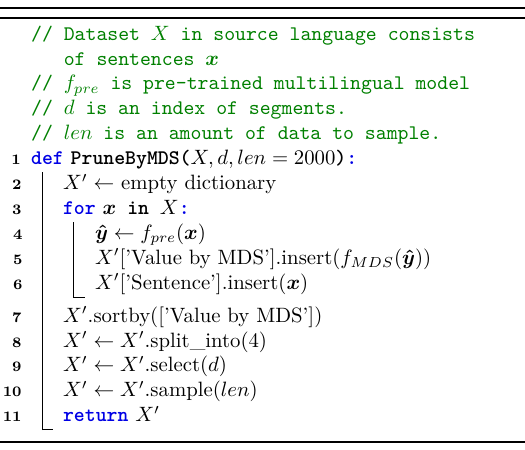}
        \caption{Pseudo code for the experiment data preparation. We sorted and split the data into 4 segments based on each value by MDS. Then, we sampled 2,000 sentences from each segment for fine-tuning.}
        \label{fig:algorithm}
    \end{center}
\end{figure}

We attempted to evaluate our method, which is one of the unsupervised MDSs, with various datasets. We sorted the data based on the values of each MDS and divided it into four segments to verify that each MDS is proportional to training-efficiency. If it is proportional and invariant across data domains, it can be regarded as 'robust guidance' for unsupervised data selection. We also conducted multiple data samplings for fine-tuning to precisely assess the capabilities of MDSs. This process follows the same cycle as described in the pseudo-code, shown in Figure \ref{fig:algorithm}. Note that the highest segment index (3 in our case) represents data subsets with the highest values according to each MDS.

\begin{table}[t]
\begin{center}
    \begin{tabularx}{4.5cm}{l|c}
     \toprule
        Data Domain & Train / Test\\
        \midrule
        Medical & 200k / 25k\\
        Travel & 160k / 20k\\
        Law & 120k / 15k\\
        Sports & 160k / 20k\\
    \bottomrule
    \end{tabularx}
    \caption{The number of sentences of 'Korean-English Parallel Corpus of Specialized Domains' dataset, released with train/test splits.}
    \label{table:amount}
\end{center}
\end{table}

\textbf{Models and Datasets} 
As a pre-trained translation model, we used 'NLLB-1.3B\citep{nllb2022}\footnote{\url{https://huggingface.co/facebook/nllb-200-distilled-1.3B}}' multilingual model. We then employed the 'Korean-English Parallel Corpus of Specialized Domains\citeplanguageresource{aihub2021}
\footnote{This research (paper) used datasets from 'The Open AI Dataset Project (AI-Hub, S. Korea)'. All data information can be accessed through 'AI-Hub (\url{www.aihub.or.kr})'.}', published by the National Information Society Agency of South Korea, as the domain-specific dataset. Given the scarcity of open datasets in the Korean language available for public download, we adopted this approach despite its limited access being restricted to nationals. There are 'Law, Medical, Travel, Sports' domains, showing each distribution in Table \ref{table:amount}. The 'Law' domain consists of precedents from the Supreme Court of South Korea. The 'Sports' domain includes various articles about international sports events. The other domains were compiled from domain-specific articles, thus containing names of locations (in the Travel domain) or names of medicines (in the Medical domain).

\textbf{Training and Hyperparameters}
Given the potential variability in domain-specific translation, such as extremely unique domains or low-resource environments, we randomly sampled 2,000 sentences from each segment, regarding the pre-defined seeds. We employed IA3 training\citep{liu2022few} to simulate practical fine-tuning environments. For hyperparameters, we set the epoch to 10, and the batch size to 32, and searched for the best learning rate from three options [1e-2, 2e-2, 3e-2] during each fine-tuning trial. Given that fine-tuning with a low-resource dataset might result in high variance between models, we took the average scores of three fine-tuned models, using sampled data with 3 different seeds.

\begin{table*}[t]
\begin{center}
\begin{tabularx}{11.7cm}{l|ccc}
    \toprule
    MDSs &\multicolumn{3}{c}{Average Performance} \\[0.25em]
    & BLEU & ChrF++ & COMET \\
    \midrule
    Not fine-tuned  & 21.42 & 45.57 & 76.39  \\
    Random  & 33.71 & 56.90 & 80.71  \\
    \midrule
    \textit{Supervised method} &&&\\[0.25ex]
    EL2N \citep{paul2021deep} & 34.01 & 57.25 & 80.84  \\ 
    \midrule
    \textit{Unsupervised methods} &&&\\[0.25em]
    Entropy \citep{jiao2021self} & 33.64 & 57.05 & 80.86 \\[0.25em]
    Selfsup \citep{sorscher2022beyond}* & 33.85 & 57.11 & 80.81\\[0.25em]
    Reference-Free COMET \citep{rei-etal-2021-references}* & 33.88 & \textbf{57.22} & \textbf{80.92}\\[0.25em]
    PerEnts (ours) & \textbf{34.09} & 57.19 & 80.82 \\
    \bottomrule
\end{tabularx}
\caption{Average test-set performance across 4 domains. We divided the dataset for each domain into four segments after sorting by each MDS and sampled 2,000 sentences three times from each segment. Given our conjecture that invariance across data domains is an important characteristic of an unsupervised MDS, we reported scores fine-tuned with subsets from either the highest (3) or lowest (0), denoted with an asterisk) segment. The highest scores among the unsupervised MDSs are highlighted in bold.}
\label{table:low_res}
 \end{center}
\end{table*}

\begin{figure*}[t]
    \begin{center}
        \includegraphics[scale=0.63]{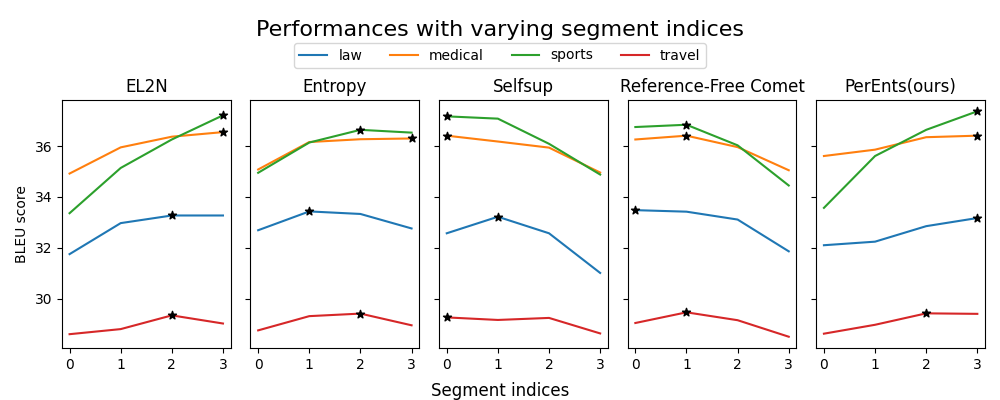} 
        \caption{The scores for each segment index across the four domains. The best BLEU scores among the segment indices were marked with a black star. Experimental results demonstrated that our method consistently identified the most training-efficient data by selecting the highest segment (3), whereas other methods varied by data domain.}
        \label{fig:segments}
    \end{center}
\end{figure*}

\begin{table*}[t]
\begin{center}
    \begin{tabularx}{13.5cm}{l|ccccc}
        \toprule
        & \multicolumn{5}{c}{The numbers of \textit{Correctly Guessed} / \textit{Newly Guessed} named entities} \\[0.5em]
        & EL2N & Entropy & Selfsup & Reference-Free & PerEnts \\
        &(Supervised)  &  &  &COMET&(Ours) \\
        \midrule
        Law  & 652/3842 & 721/3788 & 589/3837 & 690/3732 & 702/3968   \\ 
        Travel  & 2242/17389 & 2079/17693 & 1610/18159 & 2035/17686 & 1944/18554   \\ 
        Sports  & 1822/8087 & 1841/8785 & 1875/8648 & 1900/8736 & 1922/8442   \\ 
        \bottomrule
    \end{tabularx}
\caption{We observed the number of named entities that models could guess for each domain test dataset. Among the words translated by the NLLB model for each test set, named entities (NEs) were stored and classified as a 'Pre-trained Named Entities'. Additionally, NEs observed in the learning datasets created by each method were stored and classified as an 'Observed Named Entities'. If an NE inferred from a model's test data is not present in either the Pre-trained or Observed, it is categorized as 'Newly Guessed'. Furthermore, if such a guess is accurate, it is classified as 'Correctly Guessed'.}
\label{table:mem_test}
\end{center}
\end{table*}

\textbf{Implementations of MDSs}
Since our method requires named entity recognition model in the target language, which is English in our case, we employed the 'd4data/biomedical-ner-all\footnote{\url{https://huggingface.co/d4data/biomedical-ner-all}}' fine-tuned model to capture entities in the 'medical' domain dataset, such as names of medicines. For datasets in other domains, we used 'RashidNLP/NER-Deberta\footnote{\url{https://huggingface.co/RashidNLP/NER-Deberta}}' model, trained with Few-NERD dataset\citep{ding-etal-2021-nerd}, which we conjectured far more comprehensive than CoNLL-2003 dataset\citep{tjong-kim-sang-de-meulder-2003-introduction}. To implement MDS Selfsup, we used the monolingual sentence embedding model 'BM-K/KoSimCSE-roberta-multitask\footnote{\url{https://huggingface.co/BM-K/KoSimCSE-roberta-multitask}}', which is specialized for the Korean (source) language. Lastly, 'Unbabel/wmt23-cometkiwi-da-xl' was employed for Reference-Free COMET\citep{rei-etal-2021-references}\footnote{\url{https://huggingface.co/Unbabel/wmt23-cometkiwi-da-xl}}.

\subsection{Main Results}\label{sec:results}
We employed BLEU\citep{post2018call}, ChrF++\citep{popovic-2015-chrf}, and COMET scores\citep{rei-etal-2020-comet}\footnote{We used \url{https://huggingface.co/Unbabel/wmt22-comet-da} to evaluate using COMET score.} for evaluation, as presented in Table \ref{table:low_res}. The fine-tuned models were evaluated using pre-split test sets. It is important to note that, identifying the optimal value for each MDS requires access to every segment index, necessitating a complete parallel corpus for comparison. To simulate a practical strategy where access is limited, we reported averaged scores by selecting either the highest (3) or lowest (0) segment index. For instance, the domain-average score for EL2N was determined by selecting segment index 3, while for MDS Selfsup, segment index 0 was chosen.

Our method, referred to by the abbreviation 'PerEnts,' achieved the highest BLEU score among the MDSs, even surpassing the supervised method (EL2N). Although other existing methods outperformed ours for COMET and ChrF++ scores, we propose that the BLEU score might be the most critical metric for domain-specific translation due to its ability to capture semantic distortion, as demonstrated in Table \ref{table:comet-discrepancy}.

Additionally, to assess the robustness of the MDSs, we calculated the average scores across four different domains, as presented in Figure \ref{fig:segments}. The best performing segment index, selected by other MDSs, was neither 0 nor 3, suggesting that these MDSs are sensitive to the data domain. We conjectured that this observation could complement the assertion by \citet{sorscher2022beyond} that 'The best selection strategy depends on the amount of initial data.' Even though the same pre-trained weights and the same volume of data were used for each fine-tuning procedure, the data domain could play an important role as a factor. Furthermore, our selection of a well-regarded monolingual sentence embedding model\footnote{\url{https://huggingface.co/BM-K/KoSimCSE-roberta-multitask}} for implementing MDS Selfsup did not result in decent performance, supporting the idea that the sentence embedding model could be confounded by slight literal differences.

\subsection{Experiments for Generalizability}\label{sec:mem}

Fine-tuning on overly complex or specialized domains can lead to overfitting, which undermines generalization. Particularly, our method, which identifies data with complex named entities, may be prone to overfitting. To verify this, we evaluated the generalizability of each model trained with data generated by MDSs. Initially, for each test set, words translated by the NLLB model were stored and classified as a 'Pre-trained Named Entities'. Similarly, named entities identified in the training datasets selected by each MDSs were cataloged as an 'Observed Named Entities'. While translating test data, a new named entity predicted by a model, which is not in Pre-trained or Observed Named Entities, it is considered 'Newly Guessed'. If such a guess is accurate, it is deemed 'Correctly Guessed'. The counts of Newly Guessed and Correctly Guessed named entities are presented in Table \ref{table:mem_test}.

We could observed that our method do not just memorize named entities in a given train dataset. Although obvious correlations between 'Correctly Guessed Named Entities' were not exposed, our method can help a model to guess correct named entities, without an abuse generating named entities. 

\subsection{Additional Study}\label{sec:add}

\begin{table}[t]
\begin{center}
    \begin{tabularx}{7cm}{l|ccc}
     \toprule
        MDSs & \multicolumn{3}{c}{Averaged Performance} \\
        & BLEU & ChrF++ & COMET\\ 
        \midrule
        PerEnts & \textbf{34.09} & 57.16 & 80.82 \\
        \text{ *}Mean & 33.94 & \textbf{57.19} & 80.82 \\
        \midrule
        Selfsup & \textbf{33.85}&\textbf{57.11}&\textbf{80.81}\\
        \text{ *}Multilingual& 33.3&56.78&80.02\\
    \bottomrule
    \end{tabularx}
    \caption{The results of MDS variants. '*Mean' denotes that it averaged entropy instead of choosing max in our method(PerEnts), and 'Multilingual' adopted a multilingual sentence embedding model for 'Selfsup'. Both variants used the same segment index to achieve the highest average performance.}
    \label{table:ablation}
\end{center}
\end{table}

Since the intuition for each MDS could be implemented in various forms, we implemented some MDS variants. e.g., adopting average entropy instead of max for our method. We also employed multilingual sentence embedding model 'sentence-transformers/LaBSE\citep{feng-etal-2022-language}\footnote{\url{https://huggingface.co/sentence-transformers/LaBSE}}' for implementing MDS Selfsup. The results are reported in Table \ref{table:ablation}. Although there were less significant degradations, it can be argued that our method's focus on finding maximum entropy more effectively captures the 'unlearned parts.' and it reveals a limitation in the representation ability of multilingual sentence embedding models.

\section{Limitations}
We attempted to verify our method under various situations and data domains. However, it's important to note that our experiments were conducted with a single translation direction and a single data size (2k). We acknowledge that testing on multiple translation directions and diverse amounts of datasets could potentially provide a more comprehensive validation of MDSs, including our method. Additionally, the impact of utilizing named entities may vary by language, e.g., languages that use uppercase letters. Although we recognize the importance of diverse environments and theoretical analysis, limited experiments were done based on a strategic decision to verify generalizability for practical usage. We believe that these limitations could be interesting topics for future research, exploring which measurement method can generally affect the performances of fine-tuned models.

\section{Conclusion}

To identify the most training-efficient data for annotating in domain-specific machine translation, we explored various measurement methods that could serve as a benchmark for selection, collectively referred to as 'MDS.' We recognized named entities as 'complex patterns' requiring highly confident prediction. As a result, we introduced 'Capturing Perplexing Named Entity' as one of the MDSs. This approach has seen effective as a guidance for selecting training data, even in unsupervised settings. Despite the common challenge of identifying effective data for annotation in deep learning—a challenge that we could not directly address in terms of the relationship between memorizable patterns and generalizability due to a lack of theoretical analysis—we hope our findings will pave the way for more in-depth research in the future.
\nocite{*}

\section{Bibliographical References}\label{sec:reference}

\bibliographystyle{lrec-coling2024-natbib}
\bibliography{lrec-coling2024-example}

\section{Language Resource References}
\label{lr:ref}
\bibliographystylelanguageresource{lrec-coling2024-natbib}
\bibliographylanguageresource{languageresource}

\end{document}